\documentclass[pmlr]{jmlr}% new name PMLR (Proceedings of Machine Learning Research)

\usepackage{longtable}% for long tables

\usepackage{booktabs}
\usepackage[load-configurations=version-1]{siunitx} % newer version
\theorembodyfont{\upshape}
\theoremheaderfont{\scshape}
\theorempostheader{:}
\theoremsep{\newline}

\jmlrvolume{1}
\jmlryear{2022}
\jmlrworkshop{GeoMedIA Workshop 2022}

\title[Graph Neural Networks for Shape Classification in Neuroimaging]{A Comparative Study of Graph Neural Networks for\\Shape Classification in Neuroimaging}

 \author{\Name{Nairouz Shehata} \Email{n.mohamed16@imperial.ac.uk}\\
   \Name{Wulfie Bain} \Email{wulfie.bain@outlook.com}\\
   \Name{Ben Glocker} \Email{b.glocker@imperial.ac.uk}\\
   \addr Department of Computing, Imperial College London, UK}

 \editors{Jelmer Wolterink, Angelica I. Aviles-Rivero, Erik Bekkers}

\begin{document}

\maketitle

\begin{abstract}
Graph neural networks have emerged as a promising approach for the analysis of non-Euclidean data such as meshes. In medical imaging, mesh-like data plays an important role for modelling anatomical structures, and shape classification can be used in computer aided diagnosis and disease detection. However, with a plethora of options, the best architectural choices for medical shape analysis using GNNs remain unclear.\newline
We conduct a comparative analysis to provide practitioners with an overview of the current state-of-the-art in geometric deep learning for shape classification in neuroimaging. Using biological sex classification as a proof-of-concept task, we find that using FPFH as node features substantially improves GNN performance and generalisation to out-of-distribution data; we compare the performance of three alternative convolutional layers; and we reinforce the importance of data augmentation for graph based learning. We then confirm these results hold for a clinically relevant task, using the classification of Alzheimer's disease.
\end{abstract}

\section{Introduction}
\label{sec:intro}
% GDL
Geometric deep learning generalizes classical neural network models to non-Euclidean domains such as point clouds, graphs, or meshes \citep{wu2020comprehensive}. It has therefore become popular across various fields from computer vision \citep{zhou2020graph} and physics \citep{shlomi2020graph}, to healthcare topics \citep{dash2019big} such as disease prediction \citep{kazi2019graph}, drug discovery \citep{li2017learning}, and brain connectome analysis \citep{kim2021learning}. 

A recent study \citep{sarasua2022hippocampal} investigated the expressiveness of mesh representations for disease classification. We complement these findings by conducting a comparative study evaluating different graph neural networks (GNNs) for the classification of anatomical meshes extracted from neuroimaging data. We propose a simple yet effective multi-graph architecture with a shared submodel for learning shape embeddings (see Fig.~\ref{fig:pymesh}). Different graph convolutional layers are compared; GCNConv \citep{kipf2016semi}, GraphConv \citep{morris2019weisfeiler}, and SplineCNN \citep{fey2018splinecnn}. In all cases, we observe substantial performance improvements when using Fast Point Feature Histograms (FPFH) as node features, which to our knowledge has not been explored before. We also investigate the effect of data augmentation, finding improvements in generalization to data from new domains. Our findings on the proof-of-concept task of biological sex classification are confirmed on the clinically relevant diagnostic task of Alzheimer's disease classification.
\begin{figure}[htbp]
\floatconts
  {fig:pymesh}
  {\caption{Proposed multi-graph architecture; N is the number of meshes (here, N=15), H is the number of hidden features (H = 32), and FC is a fully connected layer.}}
  {\includegraphics[width=0.9\linewidth, trim = 0cm 6cm 0cm 0cm, clip]{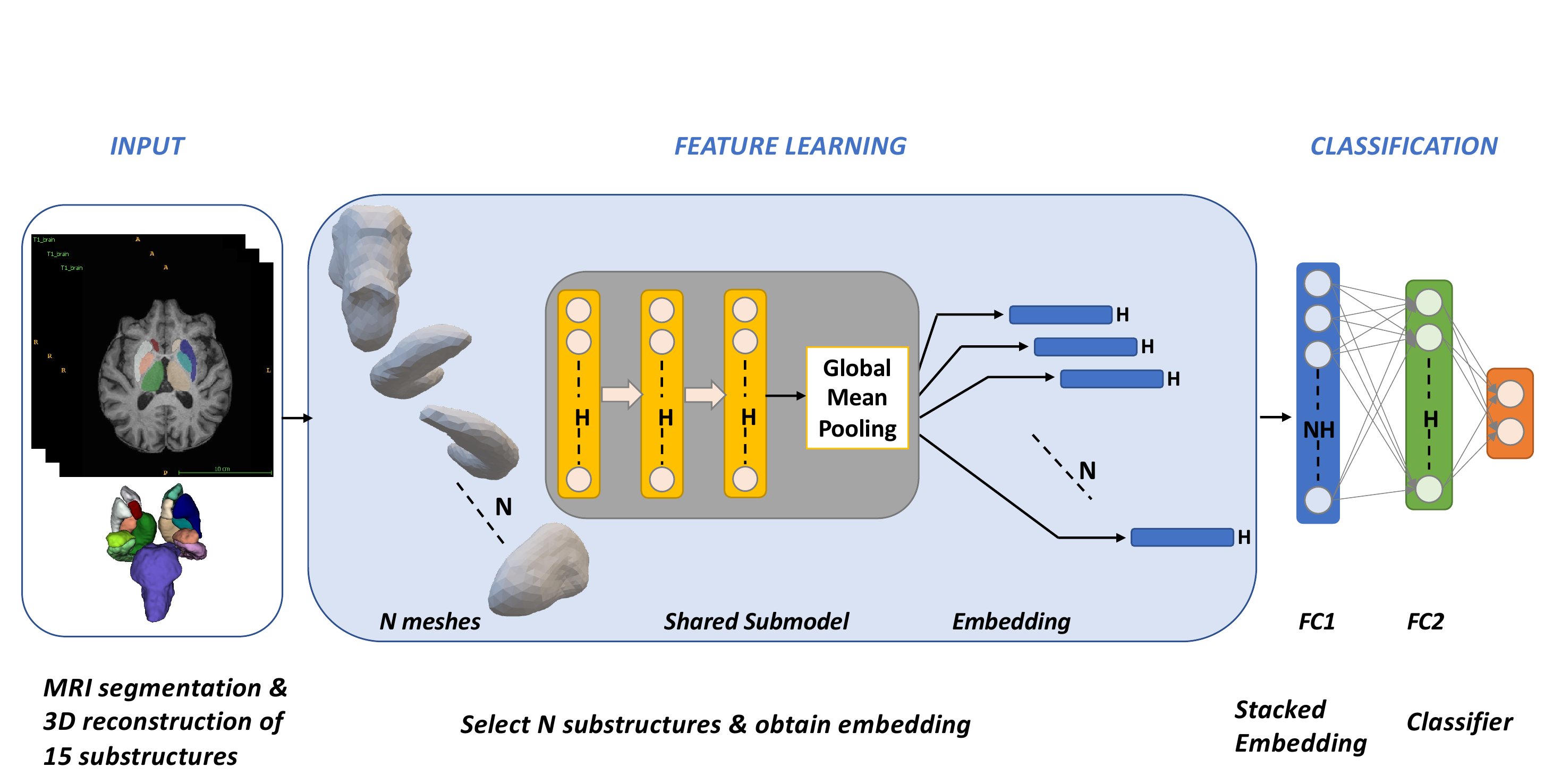}}
\end{figure}

\section{Graph Neural Network Architecture}
As the field of geometric deep learning has expanded, the architectural choices available to practitioners has proliferated. Here we outline our approach on three key aspects: the type of convolutional layer used, the number of convolutional submodels, and the type of geometric features encoded at the node level.

\subsection{Selected Graph Convolutional Operators}
Graph convolutional operations are analogous to CNN operations on images, respecting the additional invariants that arise in this domain, permutation invariance being key due to the artificial ordering of nodes that arises when representing graphs. As shown in \citet{bronstein2021GeometricDeepLearning}, many GNNs follow a blueprint of `message passing' \citep{gilmer2017neural}, whereby node features are updated using an aggregation on the features of nodes in their neighbourhood, but there is significant variance in how this is done. In this paper, we compare three seminal graph convolutional layers from the literature: GCNConv \citep{kipf2016semi}, GraphConv \citep{morris2019weisfeiler}, and SplineCNN \citep{fey2018splinecnn}. These are selected as popular representatives of graph convolutional layers, that are easy to use as plugin replacements in generic architectures. We direct readers to the original papers for details. Existing literature has compared GCNConv and GraphConv \citep{xu2018powerful,morris2019weisfeiler}, and we extend this to medical imaging.

\subsection{Multi-graph Architecture}
As multiple subcortical structure subgraphs may be extracted simultaneously from a single sample brain scan, one must also choose how to utilise these. One option is to combine them into a single multigraph per sample \citep{wang2021multi,chaari2022multigraph}. However, it might not be obvious how to define edges between graphs of different anatomical structures.\newline
Alternatively, as in this paper, each subgraph can be input to a specific GNN, and the results combined into a sample level output. Practitioners must decide the number of GNNs to use. One approach is a single shared GNN that learns from all subgraphs, while another is inputting each subgraph to a separate GNN i.e. the number of (sub) GNNs is equal to the number of subgraphs per sample \citep{hong2021spatiotemporal}. The latter approach allows each (sub) GNN to learn structure specific embeddings, whilst the former encourages the GNN to generalise learnings across structures.

Initially, we tested both a single shared and structure specific GNN submodel, finding that the performance was comparable. Using a shared submodel significantly reduces the number of parameters. Given considerations on neural networks training time \citep{li_2020}, cost \citep{wiggers_2020}, and environmental impact \citep{strubell2019energy}, our preliminary results led us to use a shared GNN in this paper: each brain substructure is passed to the shared submodel to obtain an embedding. We use three convolutional layers in the submodel with ReLU activations. A global average pooling layer is used as a readout layer to aggregate the node representations into one graph embedding. These embeddings are then stacked and passed through a fully connected layer for final classification (cf. Fig. \ref{fig:pymesh}).

\subsection{Node and edge representation}
\label{sec:graph}
The meshes representing anatomical brain structures are defined by a set of nodes and edges, where both can carry additional information. Nodes can encode arbitrary feature vectors, from spatial information such as mesh coordinates to more complex, geometric feature descriptors. In computer vision, hand crafted features based on carefully designed descriptors have been largely abandoned in the end-to-end deep learning paradigm \citep{battaglia2018relationalinductivebiases}. However, in the case of shape analysis, we believe there is value in sophisticated, geometrical feature extractors, especially when there are limited amounts of training data. We evaluate the use Fast Point Feature Histograms (FPFH) \citep{rusu2009fast} as node features, and compare these with positional node features in form of Cartesian coordinates, and no node features (realized by setting constant values).

To calculate the FPFH features on a mesh, first a point feature histogram is computed: for each query point $p_r$, all neighbouring points inside a 3D sphere of radius $r$ centered at point $p_r$ are selected (k-neighbourhood points); then, for each pair \(p_r\) and \(p_k\) in the k-neighbourhood points of $p_r$, their normals are estimated as $n_r$ and $n_k$. The point with the smaller angle between the line joining the pair of points and the estimated normals is chosen to be $p_r$. Finally a \textit{Darboux frame} is defined as (\(u = n_r, v = (p_k - p_r) × u, w = u \times v\)) and the angular variations of \(n_r\) and \(n_k\) are computed:
\[\alpha = v \cdot n_k\]
\[\phi = (u \cdot (p_k - p_r))/\left\|(p_k - p_r)\right\|\]
\[\theta = arctan(w \cdot n_k , u \cdot n_k )\]
Second, a Simple Point Feature Histogram (SPFH) is obtained by calculating the point features of each neighboring point $p_k$ \citep{rusu2008learning}.
Finally, to calculate FPFH, the SPFH of the $k$ neighbours are used to calculate the final histogram of $p_r$, where they are weighted by the distances between $p_r$ and the neighbours $p_k$. $N$ is the number of points within the sampling radius (number of neighbours to the reference point). In our implementation, the sampling radius was set to 10mm and maximum number of neighbours to 100.

\[FPFH(p_r) = SPFH(p_r) + \frac{1}{N} \sum_{k=1}^N \frac{SPFH(p_k)}{\left\| p_k - p_r\right\|}\]

Besides the node features, we also encode edge attributes in terms of relative spherical coordinates between two nodes. Edge attributes are processed only within SplineCNN layers, but otherwise ignored in both GCNConv and GraphConv layers, as these can only use edge weights and not attributes.

\section{Datasets}
We utilize four neuroimaging datasets to test generalization and robustness of the classification performance. We use data from the UK Biobank imaging study (UKBB)\footnote{UK Biobank Resource under Application Number 12579} \citep{sudlow2015uk,miller2016}, the Cambridge Centre for Ageing and Neuroscience study (Cam-CAN) \citep{shafto2014cambridge,taylor2017cambridge}, and the IXI dataset\footnote{\url{https://brain-development.org/ixi-dataset/}}. Both UKBB and Cam-CAN use a similar imaging protocol with Siemens 3T scanners. IXI consists of data acquired at three different sites including Guy’s Hospital using a Philips 1.5T system, Hammersmith Hospital using a Philips 3T scanner, and Institute of Psychiatry using a GE 1.5T system. UKBB, Cam-CAN, and IXI are data from healthy volunteers. We only discarded data related to subjects whose sex or age entries were unavailable. 

We also use the OASIS-3 dataset with 716 cognitively normal participants and 318 participants who reach various stages of cognitive decline during the study, allowing Alzheimer’s disease (AD) related tasks such as classification \citep{lamontagne2019oasis}. The cognitive status is reflected in the clinical dementia rating (CDR) that accompanies the imaging dataset, with subjects receiving a score of: 0 for normal, 0.5 for very mild dementia, 1 for mild dementia, 2 for moderate dementia and 3 for severe dementia \citep{morris1991clinical}. The CDR is collected in clinical sessions, separate to the imaging sessions, meaning sessions must be `matched' to get an \{image, CDR score\} pair. We match the clinical diagnosis closest in time to each scan, before filtering out samples where the absolute time difference between scan and clinical assessment is greater than 365 days. To avoid difficulties in assigning scans to training, validation, and testing, we only use one scan per subject, leaving 1,084 unique scans. We exclude 50 samples because their sex or age information was missing. The final set of 1,034 comprises 716, 188, 111, 18 and 1 samples, for CDR of 0, 0.5, 1, 2 and 3 respectively. We binarize CDR to 0 and 1 (for CDR score 0.5, 1, 2 and 3).

The UKBB data comes pre-processed with already extracted meshes for 15 subcortical brain structures\footnote{Brain stem, left/right thalamus, caudate, putamen, pallidum, hippocampus, amygdala, accumbens-area}. We apply our own processing pipeline to Cam-CAN, IXI and OASIS-3 to match UKBB as closely as possible: 1) Skull stripping with ROBEX v1.2\footnote{\url{https://www.nitrc.org/projects/robex}} \citep{iglesias2011robust}; 2) Bias field correction with N4ITK\footnote{\url{https://itk.org}} \citep{tustison2010n4itk}; 3) Sub-cortical brain structure segmentation and meshing using FSL FIRST \footnote{\url{https://fsl.fmrib.ox.ac.uk/fsl/fslwiki/FIRST}} \citep{patenaude2011bayesian}.

\begin{table}[hbtp]
\floatconts
  {tab:data}
  {\caption{Number of samples, percentage of females, and mean, min, and max age.}}
  {\begin{tabular}{lccc}
  \toprule
  \bfseries Dataset & \bfseries Samples & \bfseries Female (\%) & \bfseries Age (years)\\
  \midrule
  UKBB & 13,749 & 47 & 61 [44, 73]\\
  Cam-CAN & 652 & 51 & 54 [18, 88]\\
  IXI & 563 & 58 & 49 [20, 86]\\
  OASIS-3 & 1,034 & 55 & 72 [42, 97]\\
  \bottomrule
  \end{tabular}}
\end{table}

\section{Experiments}
\label{sec:experiments}
The experiments were designed to evaluate and compare three main aspects: (i) the choice of convolutional layers for the shared submodel; (ii) the choice for the node features; (iii) the effect of data augmentation on robustness and generalization.

\subsection{Implementation and Training}
We use the Adam optimizer with a learning rate of 0.001 and the standard cross entropy loss as the classification objective function. To increase the variability of the training data and to avoid overfitting, we employ a simple data augmentation strategy \citep{zhou2020data}. Individual graph nodes are randomly translated by a maximum offset. We evaluate the effect of the strength of augmentation and test maximum offsets of 0.1mm, 0.5mm, and 1.0mm. Given the limited amount of training data, data augmentation should be beneficial for improving classification accuracy across different datasets.

All our implementations were done in PyTorch benefiting from the excellent PyTorch Geometric library\footnote{\url{https://pytorch-geometric.readthedocs.io/}}. We use PyTorch Lightning\footnote{\url{https://www.pytorchlightning.ai/}} for ease of implementation of the model and data structures. The code is available on \url{https://github.com/biomedia-mira/medmesh}.

\subsection{Task 1: Biological Sex Classification}
\label{task1}
We use biological sex classification as a proof of concept task which has shown to yield good performance with the advantage that several neuroimaging datasets from different sources are available for extensive testing and evaluation of the effect of different model choices on predictive performance. We use the UKBB data for the model development, with a data split of 70\%, 10\%, and 20\% for training, validation, and testing. The batch size was set to 128, all hidden features set to 32 (both for the convolutional layers and fully connected layers). When using SplineCNN, we set the kernel size to 5 and use the sum aggregation. The maximum number of training epochs was set to 50, and we retain the model with highest validation performance for final evaluation on the test set.

\paragraph{\textbf{Effect of node features}}
To evaluate the effectiveness of different node features, in our first set of experiments we employ SplineCNN in the shared convolutional submodel (as these performed well in initial experimentation). We then evaluated classification performance on the UKBB test set, Cam-CAN, IXI, and OASIS-3 using constant, positional, and FPFH node features.

The ROC curves in Figure \ref{fig:sex-feat} show that FPFH substantially outperforms other node features on all four datasets. It is worth noting that while positional features perform well on the in-distribution UKBB test set, these features underperform on out-of-distribution test sets. This is due to their reliance on Cartesian coordinates of mesh nodes which do not generalize well due to differences in data acquisition. FPFH, on the other hand, are invariant to the pose of the mesh and show much better generalization across datasets.

\begin{figure}[htbp]
\floatconts
  {fig:sex-feat}
  {\caption{ROC curves for sex classification comparing different node features across the datasets UKBB, Cam-CAN, IXI, OASIS-3 using SplineCNN in the submodel.}}
  {%
    \subfigure[UKBB]{\label{fig:image-ukbb}%
      \includegraphics[width=0.35\linewidth]{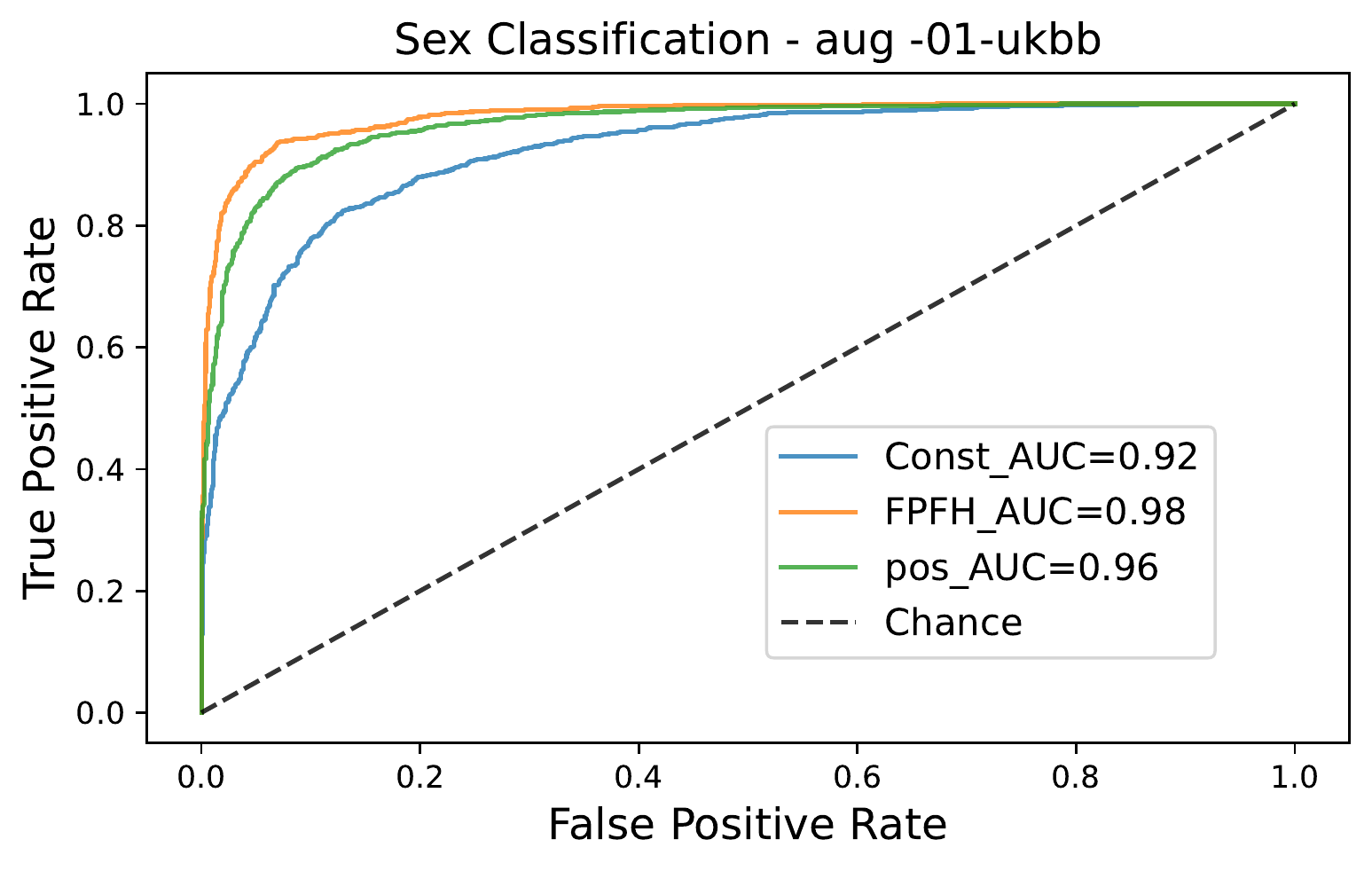}}
    \subfigure[Cam-CAN]{\label{fig:image-camcan}%
      \includegraphics[width=0.35\linewidth]{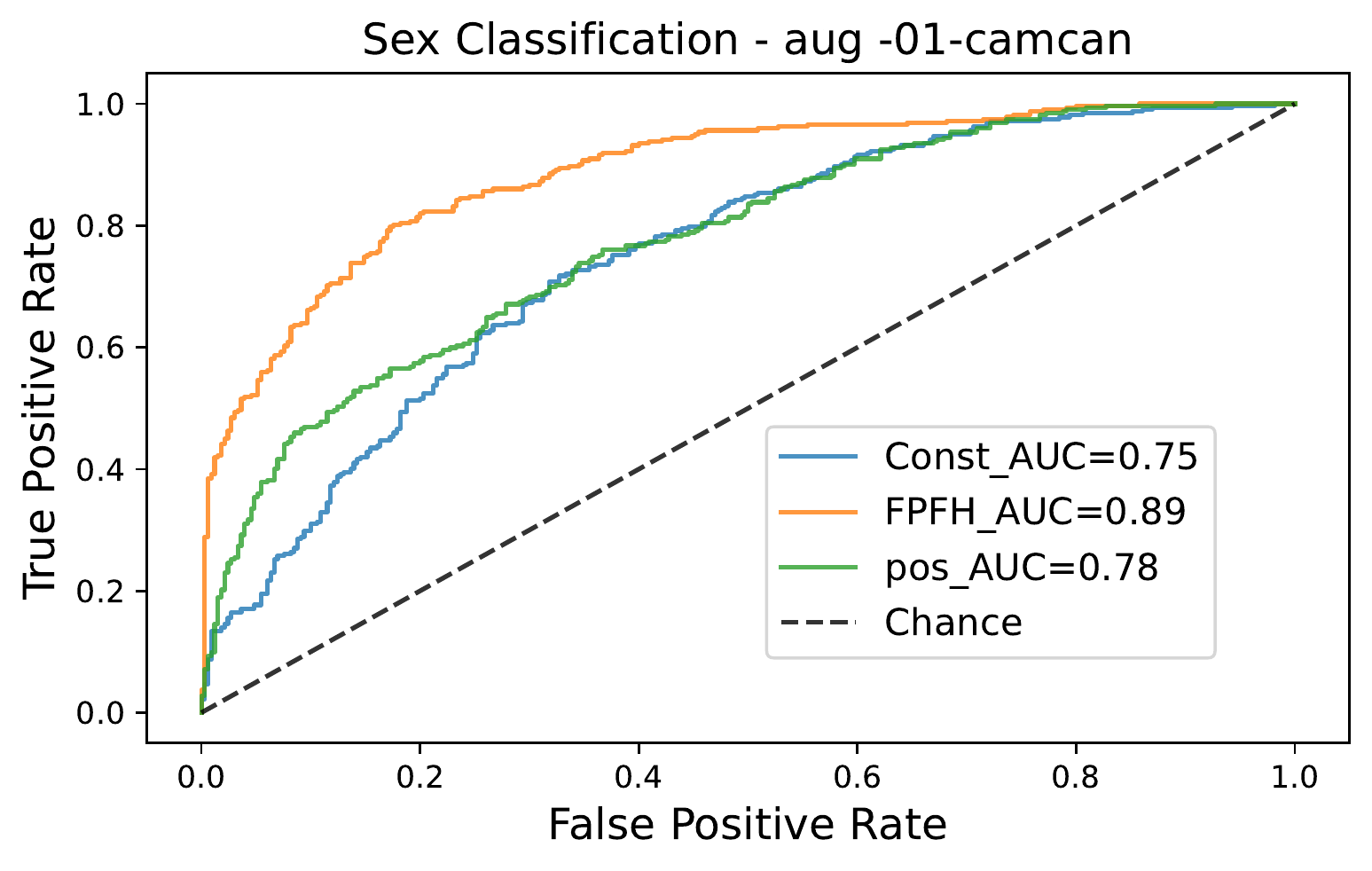}}
    \subfigure[IXI]{\label{fig:image-ixi}%
      \includegraphics[width=0.35\linewidth]{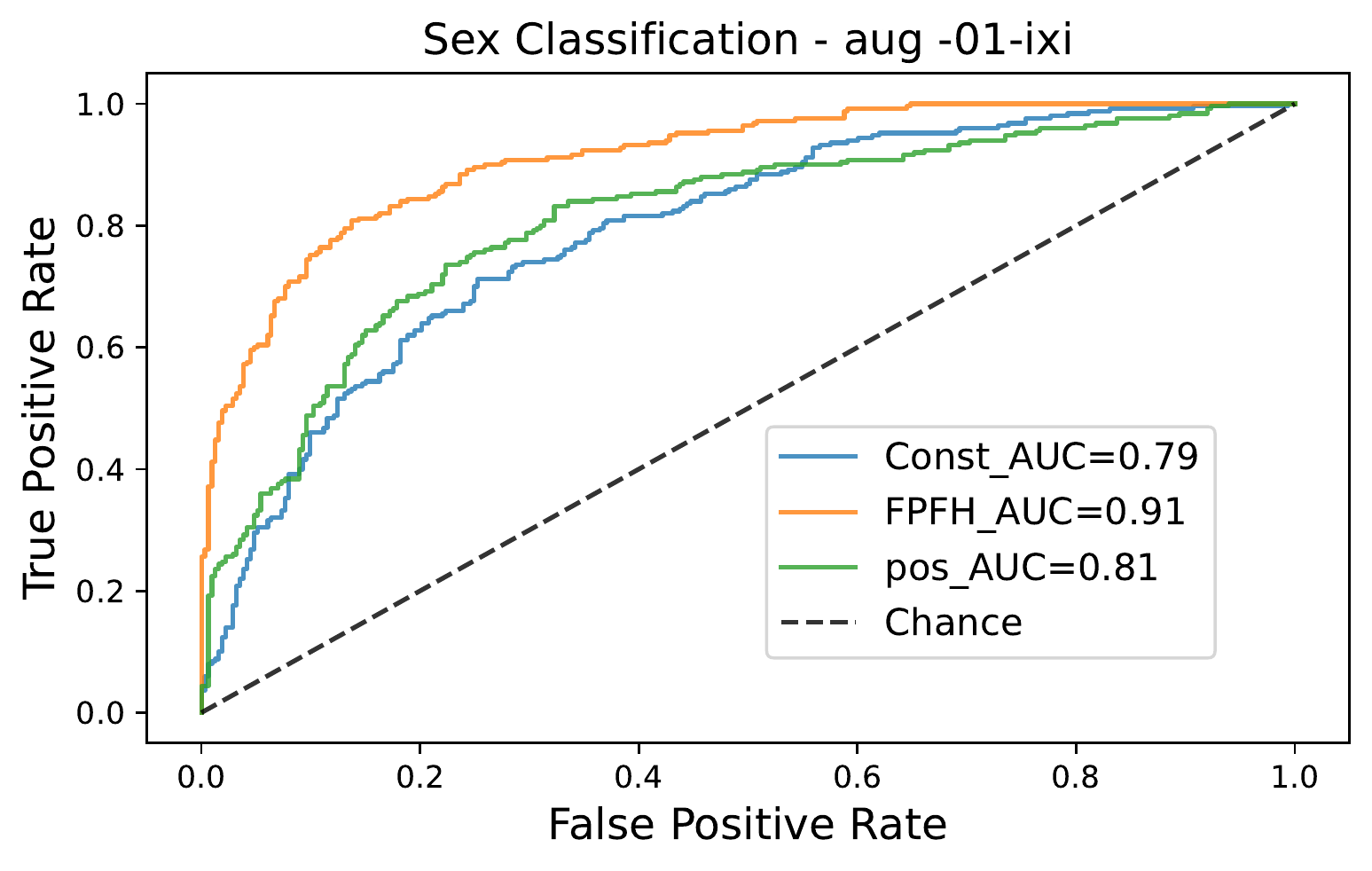}}
    \subfigure[OASIS-3]{\label{fig:image-oasis3}%
      \includegraphics[width=0.35\linewidth]{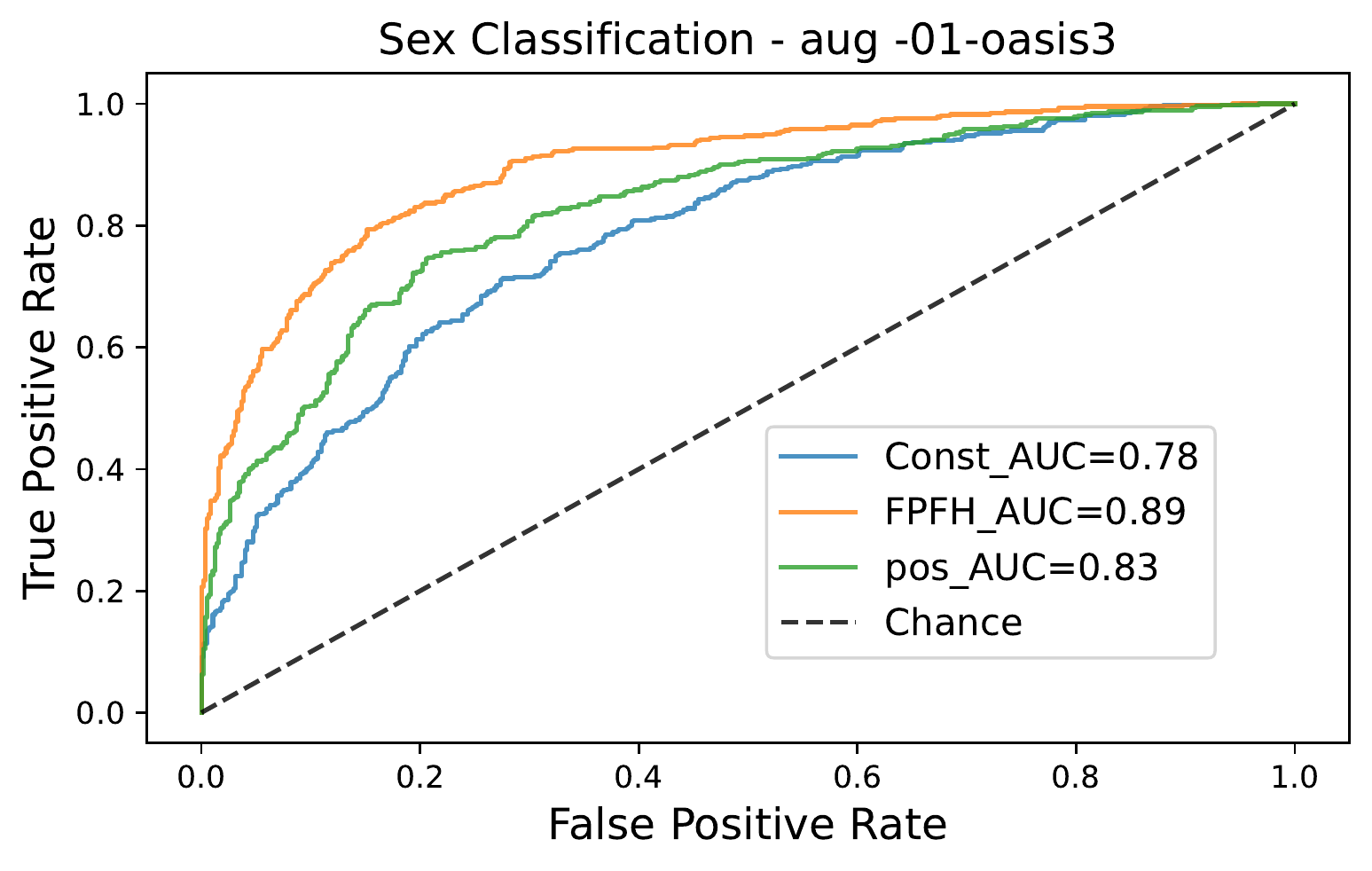}} 
  }
\end{figure}

\paragraph{\textbf{Effect of data augmentation}}
Next, we evaluate the effect of varying strengths of data augmentation. The maximum offset for the random node translation is varied from 0 (no augmentation), to 0.1, 0.5, and 1.0mm. The ROC curves in Figure \ref{fig:sex-aug} demonstrate the benefit of data augmentation on robustness and generalization. The best performance is achieved using data augmentation of 0.1, which increased AUC by 4-5\% compared to not using augmentation. While data augmentation slightly decreases the performance on the in-distribution UKBB test set, it substantially improves performance on all out-of-distribution test sets, confirming the importance of adding random perturbations to the training data.

\begin{figure}[htbp]
\floatconts
  {fig:sex-aug}
  {\caption{ROC curves showing the effect of data augmentation for sex classification across domains, using SplineCNN as the shared submodel and FPFH as node features.}}
  {%
    \subfigure[UKBB]{\label{fig:image-fpfh-ukbb}%
      \includegraphics[width=0.35\linewidth]{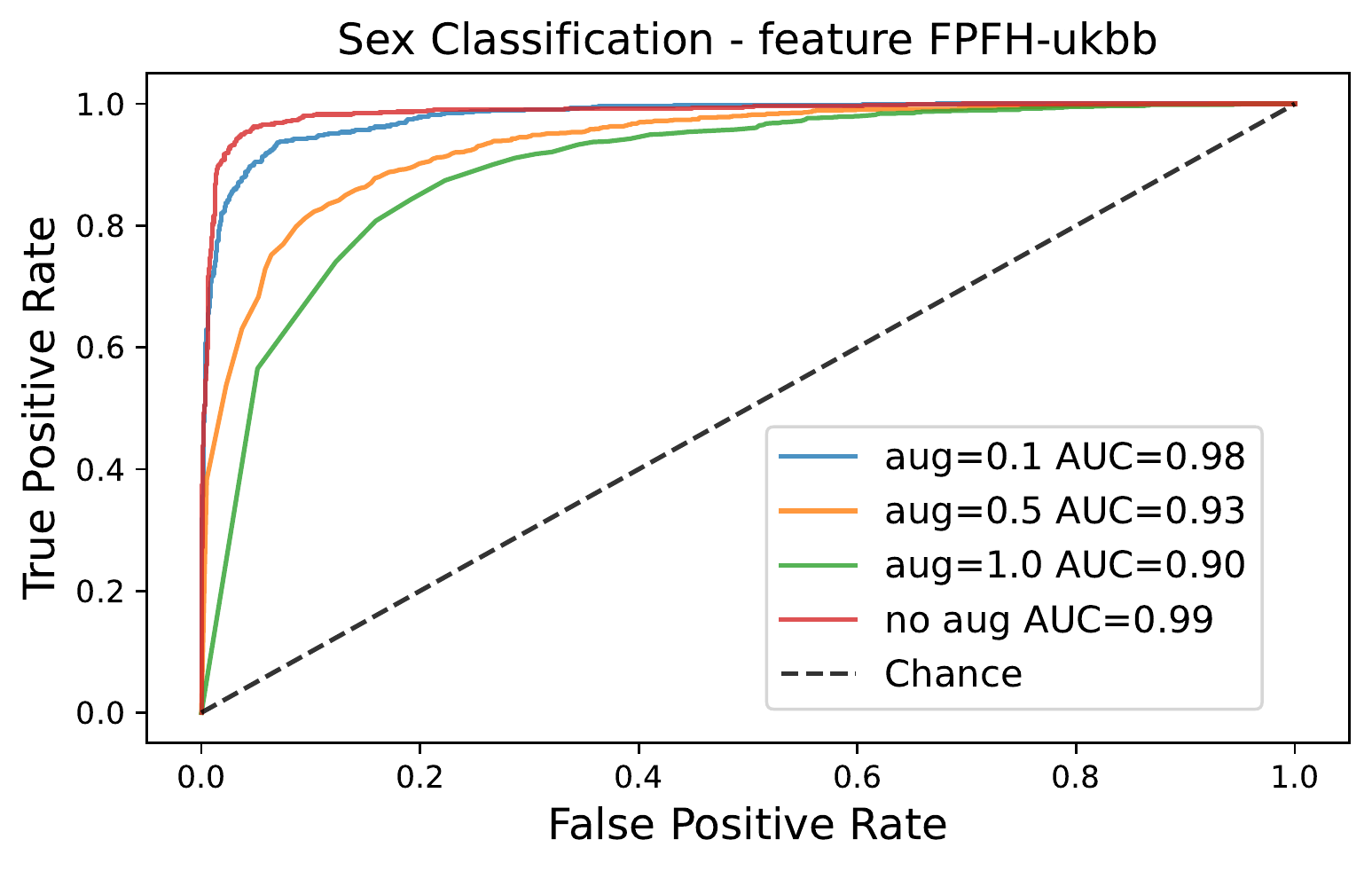}}
    \subfigure[Cam-CAN]{\label{fig:image-fpfh-camcan}%
      \includegraphics[width=0.35\linewidth]{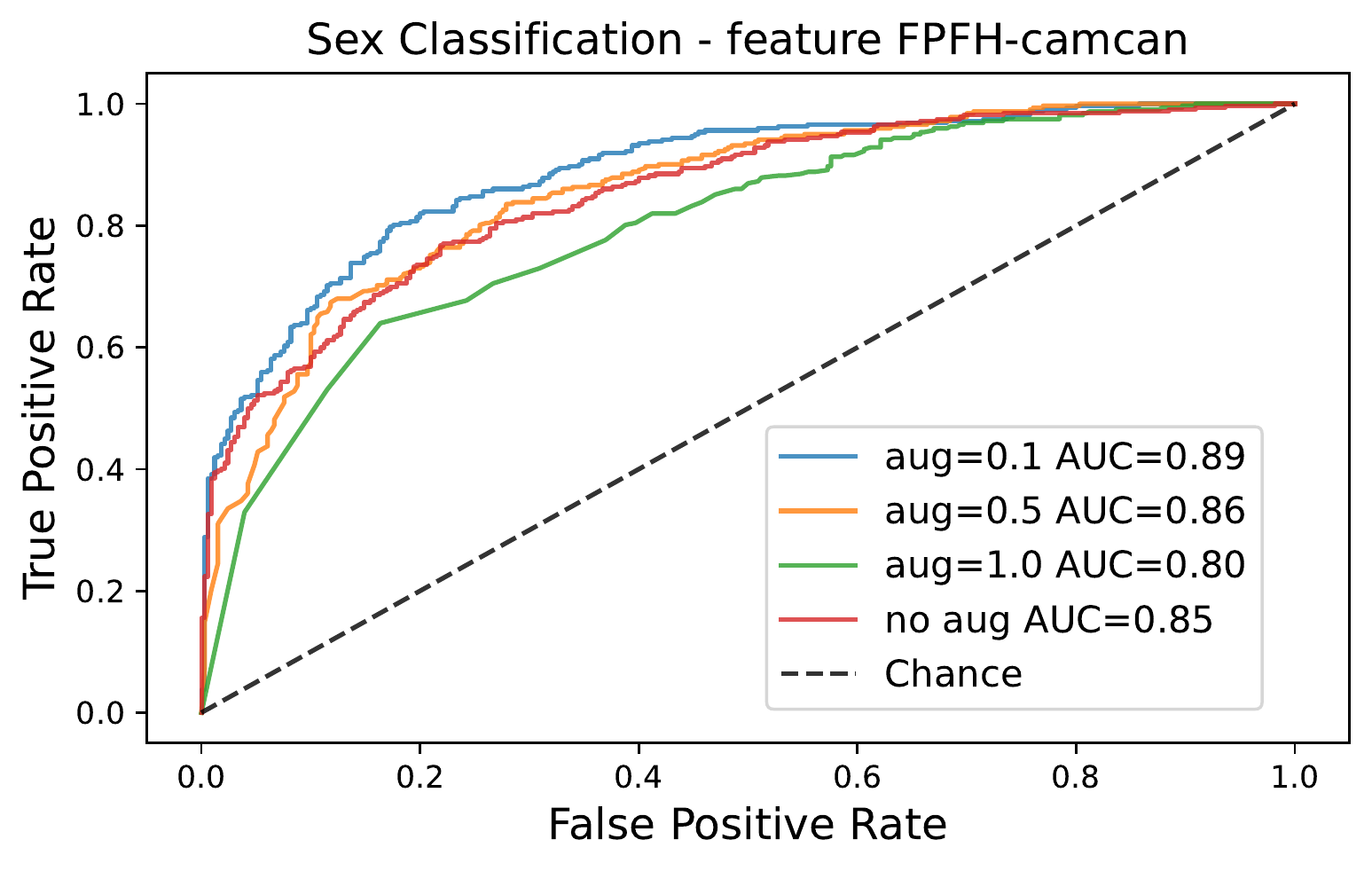}} 
    \subfigure[IXI]{\label{fig:image-fpfh-ixi}%
      \includegraphics[width=0.35\linewidth]{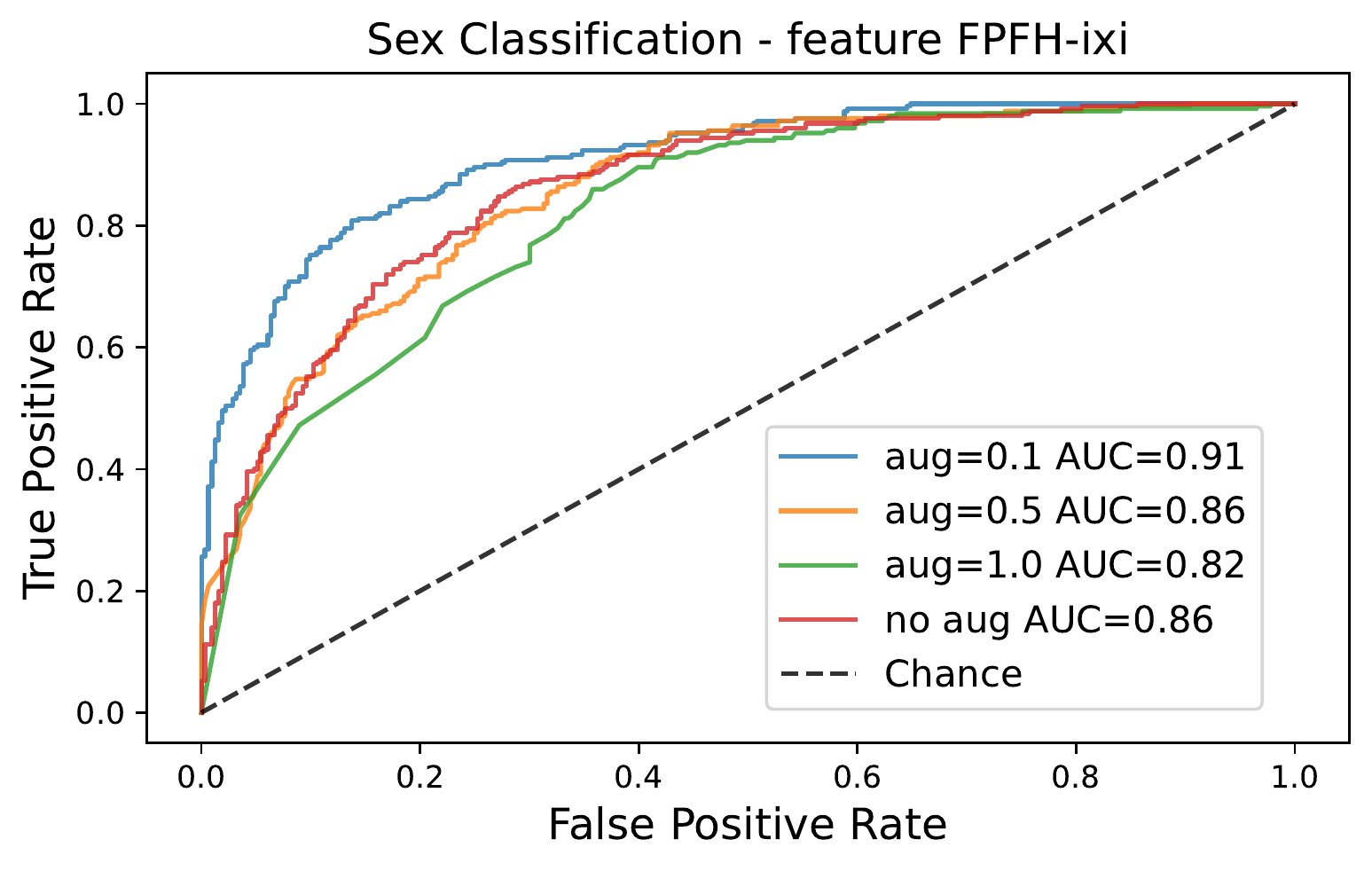}} 
    \subfigure[OASIS-3]{\label{fig:image-fpfh-ixi}%
      \includegraphics[width=0.35\linewidth]{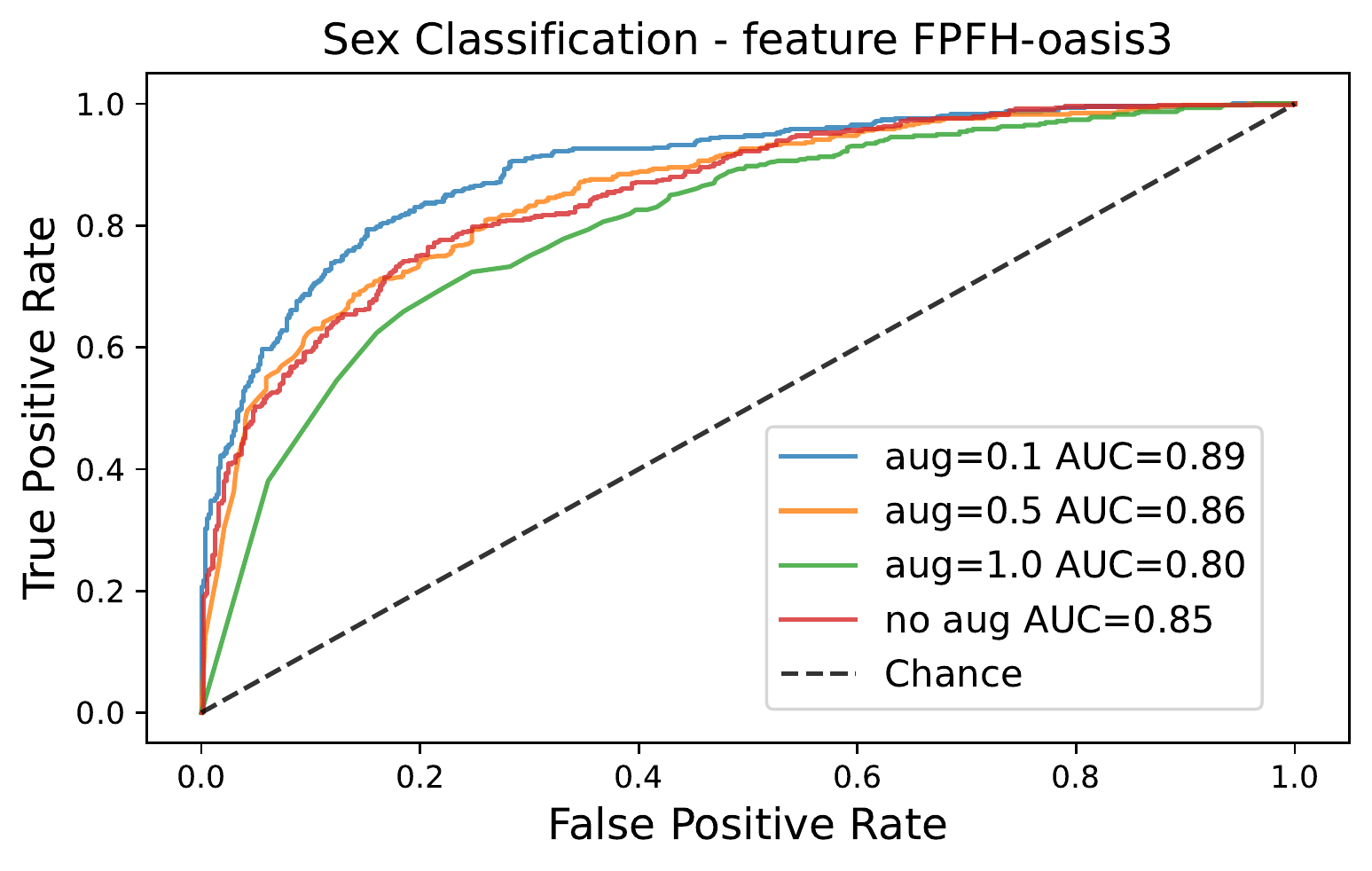}}
  }
\end{figure}

\paragraph{\textbf{Effect of convolution layer}} Finally, we evaluate the three different convolutional layers, using FPFH as the node features and data augmentation of 0.1mm. In Figure \ref{fig:sex-layer} we observe similar performance for SplineCNN and GCNConv, closely followed by GraphConv.

\subsection{Task 2: Alzheimer's Disease Classification}
\begin{figure}[htbp]
\floatconts
  {fig:layers-and-AD}
  {\caption{Effect of convolution layer for (a) sex and (b) Alzheimer's disease classification. (c) Effect of node features on AD classification. All evaluated on OASIS-3.}}
  {%    
    \subfigure[Sex - Conv. Layers]{\label{fig:sex-layer}
      \includegraphics[width=0.31\linewidth]{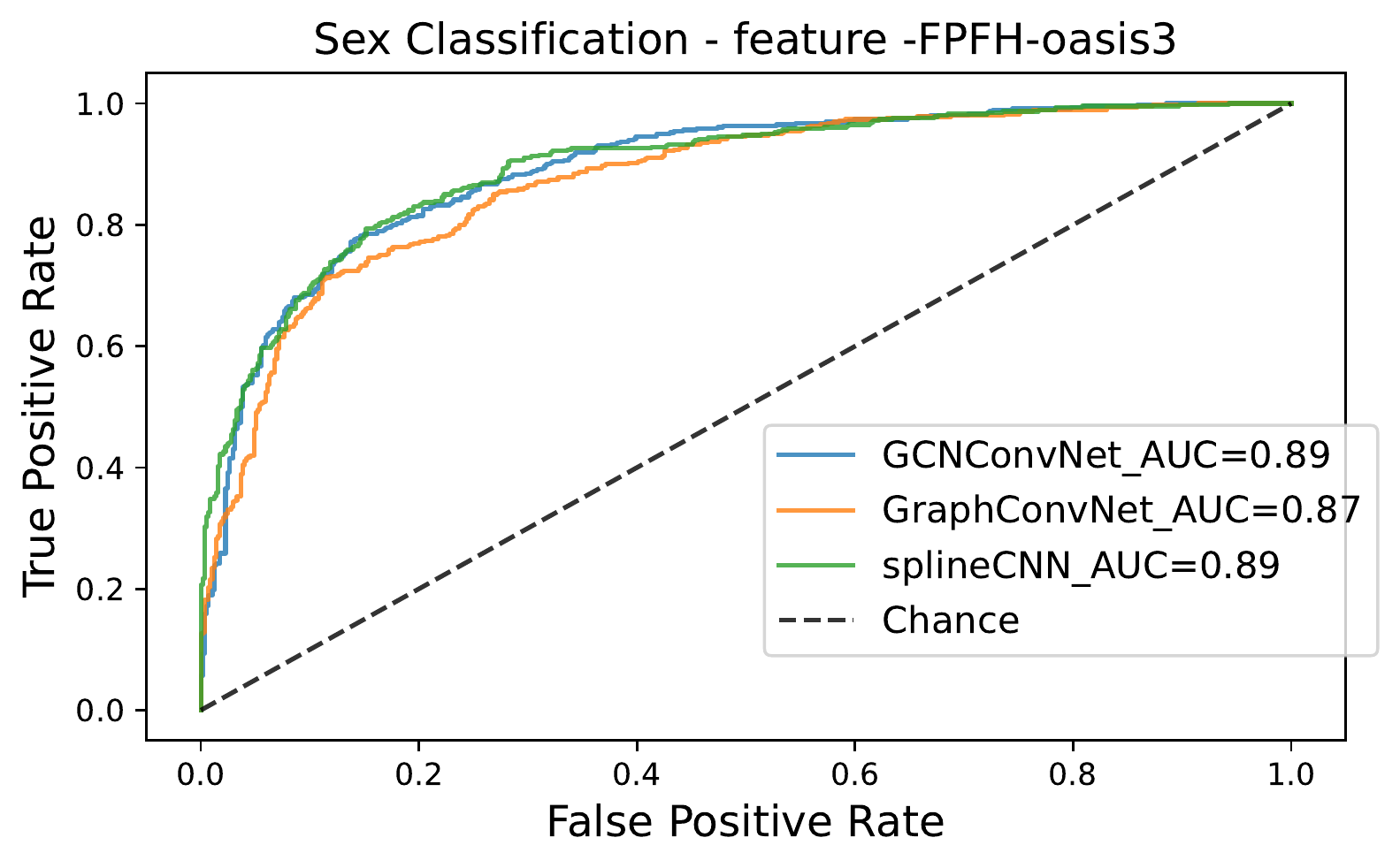}}
    \subfigure[AD - Conv. Layers]{\label{fig:ad-layer}
      \includegraphics[width=0.31\linewidth]{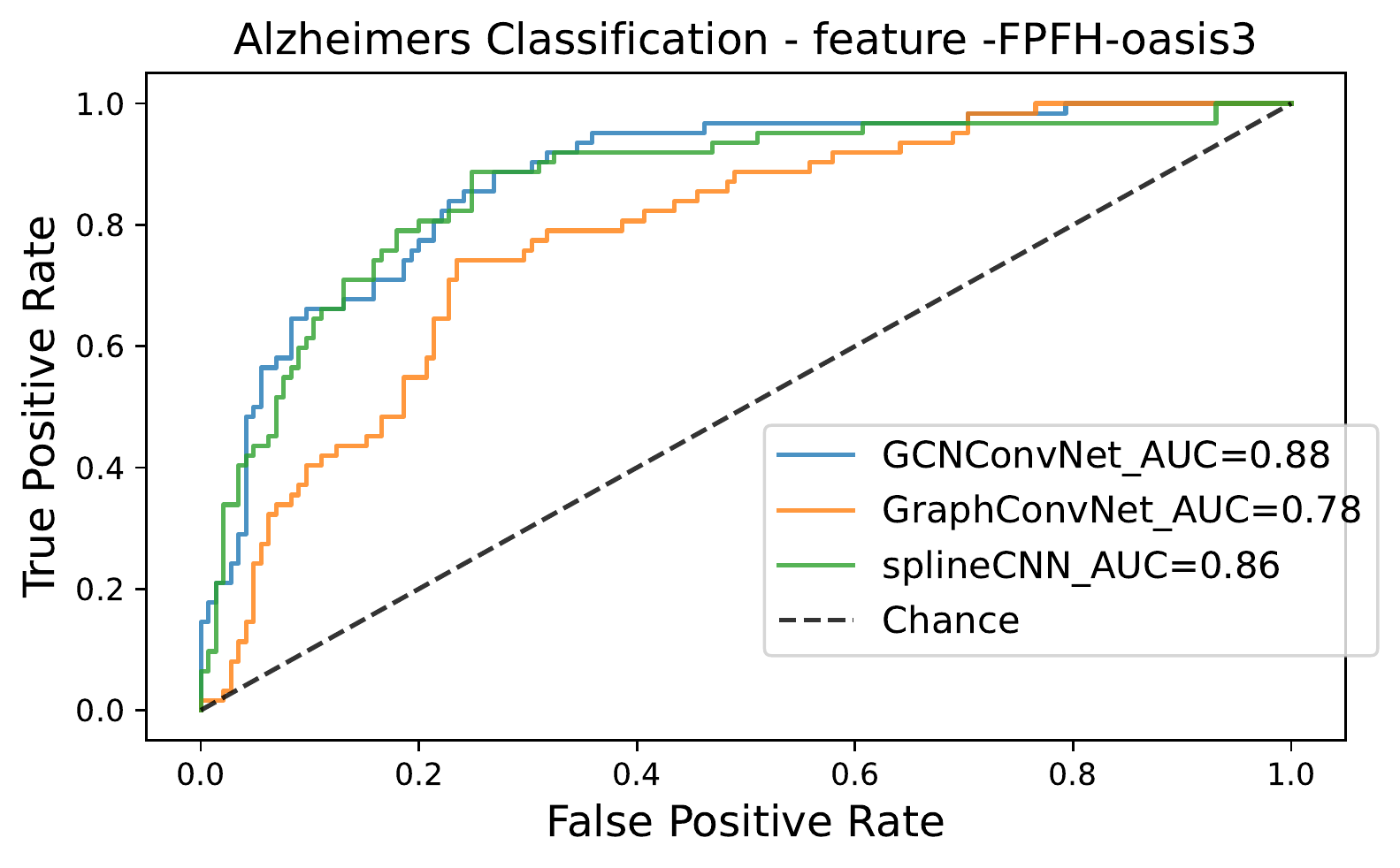}}
    \subfigure[AD - Node Features]{\label{fig:ad-feat}
      \includegraphics[width=0.31\linewidth]{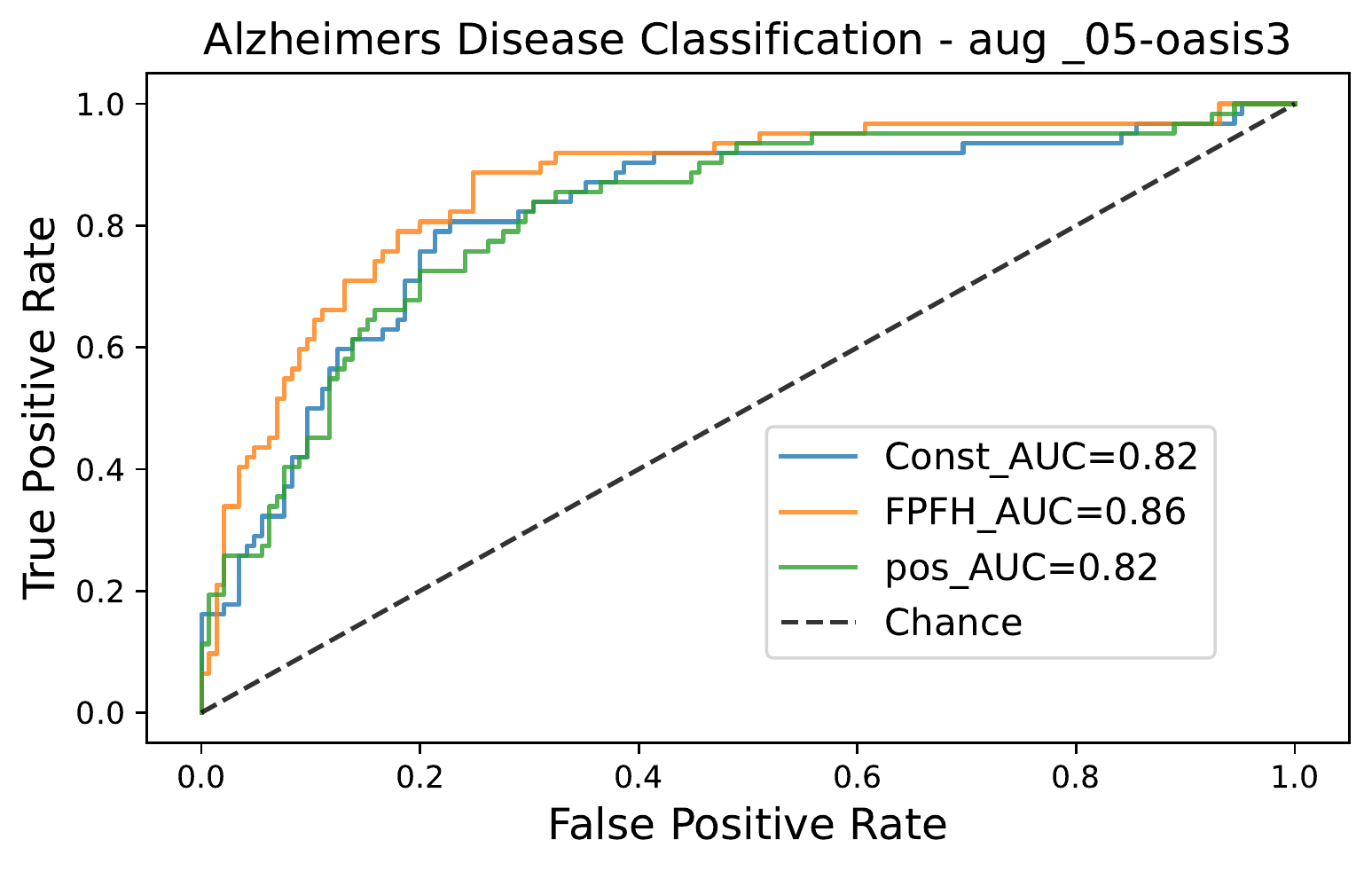}}
  }
\end{figure}

To confirm whether the above findings hold for a clinically relevant task, we consider Alzheimer's disease (AD) classification on OASIS-3 with a 70\%, 10\%, and 20\% train, validation, and test split. We evaluate the effect of the convolutional layer using a larger amount of data augmentation of 0.5mm due to the smaller amounts of training data. We then also evaluate the effect of node features for AD classification, using SplineCNN in the submodel for consistency with the sex classification experiments. The results are shown in Fig. \ref{fig:ad-layer} and \ref{fig:ad-feat}. GCNConv performs slightly better than SplineCNN, with a substantial decrease in performance for GraphConv. FPFH features again outperform other node features.

\subsection{Bias Analyses}
\begin{figure}[htbp]
\floatconts
  {fig:strat}
  {\caption{Bias analysis for sex classification using SplineCNN with FPFH features. Classification performance is stratified by (a) age groups and (b) presence of disease.}}
  {%
    \subfigure{\label{fig:sexstrat}
      \includegraphics[width=0.48\linewidth]{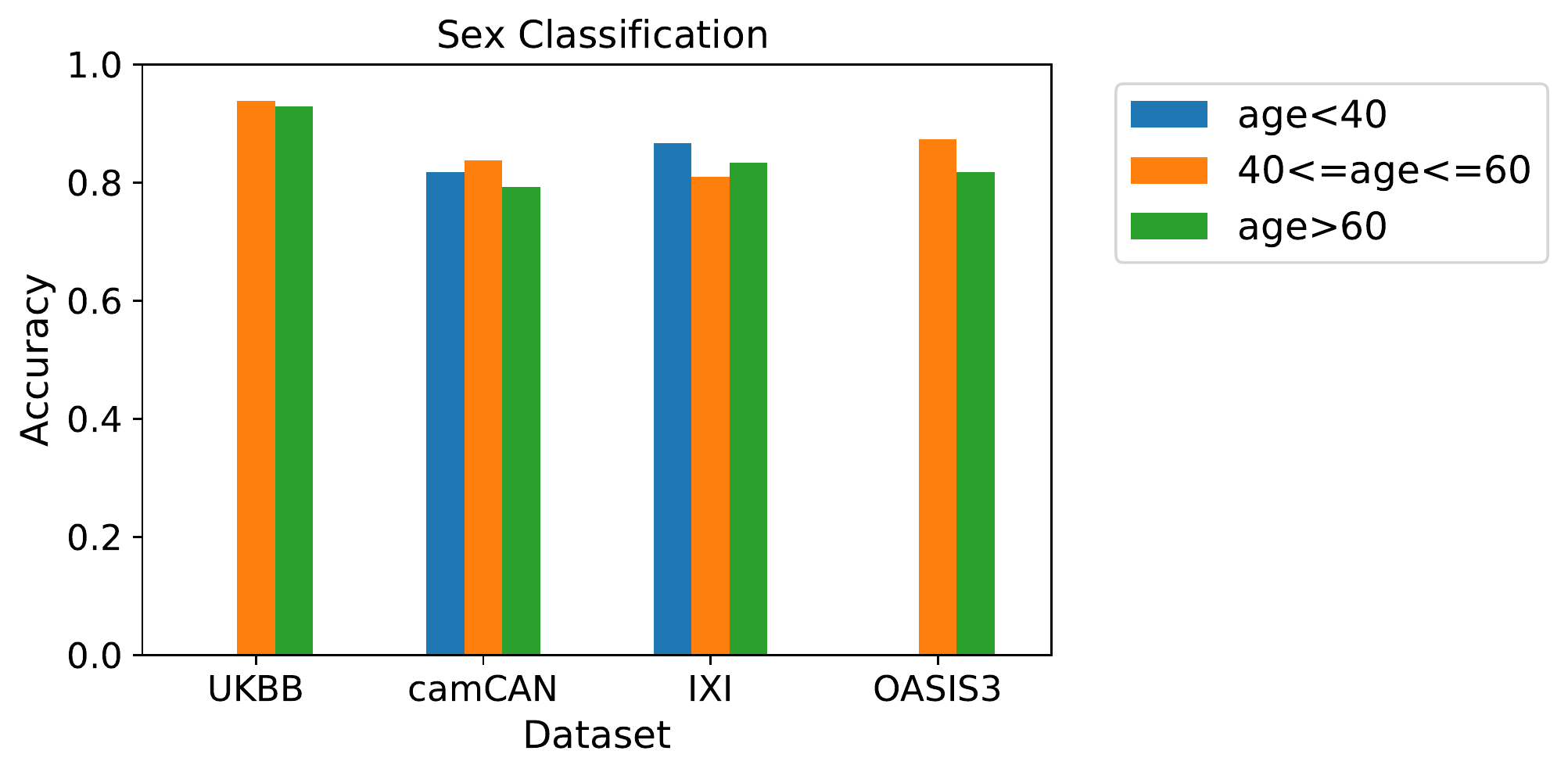}}
    \subfigure{\label{fig:adstrat}
      \includegraphics[width=0.35\linewidth]{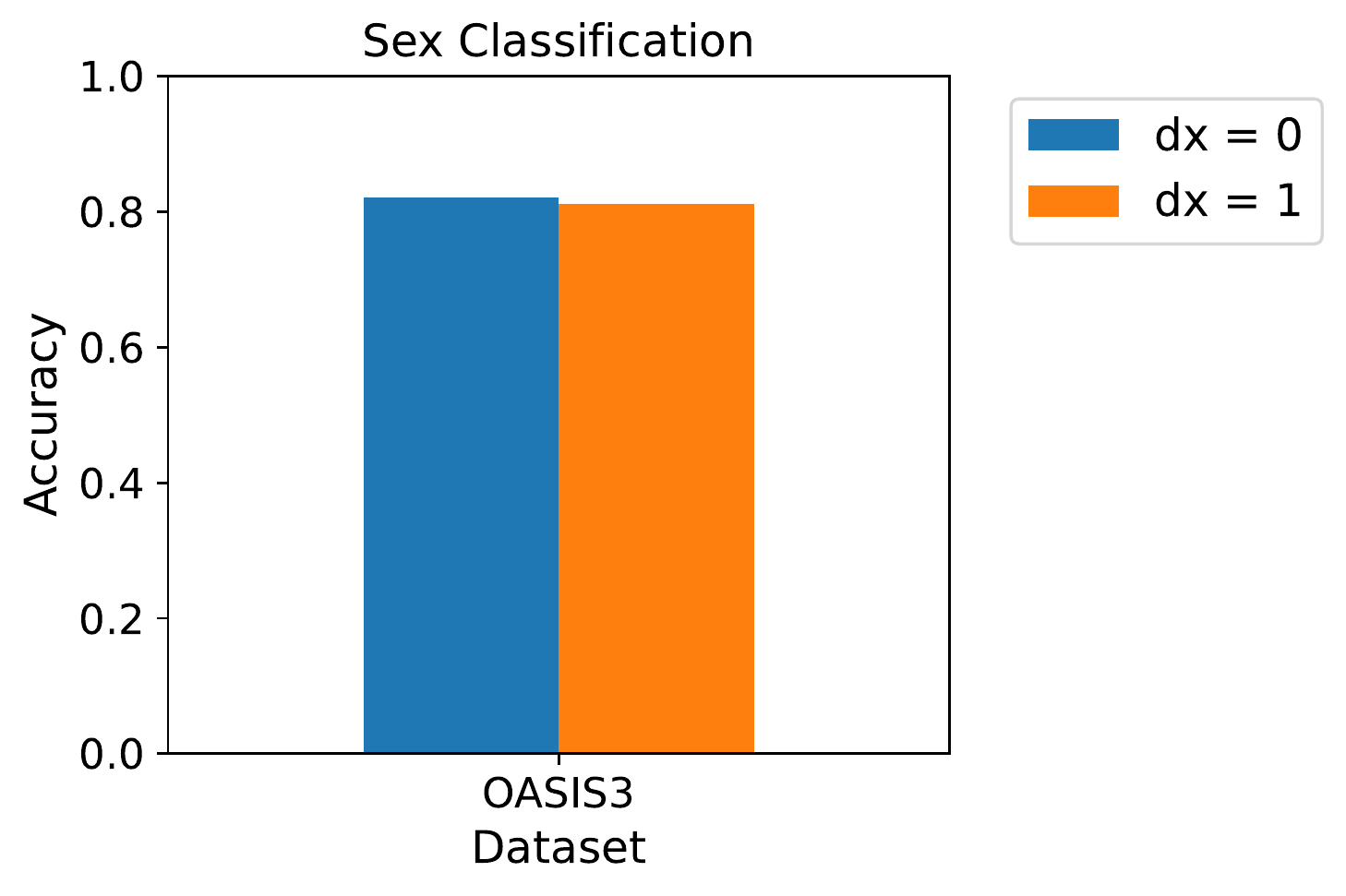}}
  }
\end{figure}

We also investigated potential biases in the predictions in terms of subgroup performance disparities. To this end, we first analyzed the biological sex classification model stratified by age groups. As the training data from UKBB only covers a limited age range between 44 and 73 year old subjects, we wanted to understand whether the performance might degrade for younger subjects. The results shown in Figure \ref{fig:sexstrat}, however, suggest that the sex classification model with SplineCNN and FPFH features generalizes well across the entire age range. Both Cam-CAN and IXI contain many subjects in the range of 18 to 40 years. Next, we analyzed whether sex classification may be affected by disease status. Here we looked at the classification performance separately for the group of healthy controls and subjects with Alzheimer's disease. Again, we find no differences in the classification accuracy, suggesting that the sex classification model generalizes well (cf. Figure \ref{fig:adstrat}).

\section{Conclusion}
\label{sec:conclusion}
This comparative study evaluated the effect of node features, convolutional layers, and data augmentation on two different tasks and four datasets in medical shape classification with graph neural networks. We find that the use of FPFH features is highly beneficial, substantially improving classification performance on out-of-distribution test data. The FPFH features alleviate the need for data normalization such as mesh alignment due to their pose invariance. We are not aware of earlier studies proposing the use of FPFH in GNNs. We further find that SplineCNN and GCNConv are both viable options for the convolutional layers, yielding comparable performance. We also conclude that data augmentation is essential in GNNs, in particular, when the amount of training data is limited. We find that stronger data augmentation is beneficial in particular for Alzheimer's disease classification where the training set contained less than 800 samples. AD classification performance was overall promising, and in line with a recent study evaluating data representations \citep{sarasua2022hippocampal}. This should be confirmed in future work on other datasets for AD classification such as ADNI.

A limitation of this work is that only relatively simple data augmentation was considered in the form of perturbing the mesh node positions. Recently, there has been work on more advanced data augmentation techniques for graph neural networks \citep{ding2022data} such as removing a certain number of edges either randomly \citep{rong2019dropedge} or based on the GNN predictions adjusting the graph in an adaptive manner \citep{Chen_Lin_Li_Li_Zhou_Sun_2020} to name a few. In future work, it would be interesting to evaluate the effect of these more advanced techniques for medical shape classification.

Our proposed multi-graph architecture may be useful in other applications which will be investigated in future work.

\acks{Nairouz Shehata is grateful for the support by the Magdi Yacoub Heart Foundation.}

\bibliography{shehata22}

\begin{thebibliography}{36}
\providecommand{\natexlab}[1]{#1}
\providecommand{\url}[1]{\texttt{#1}}
\expandafter\ifx\csname urlstyle\endcsname\relax
  \providecommand{\doi}[1]{doi: #1}\else
  \providecommand{\doi}{doi: \begingroup \urlstyle{rm}\Url}\fi

\bibitem[Battaglia et~al.(2018)Battaglia, Hamrick, Bapst, Sanchez{-}Gonzalez,
  Zambaldi, Malinowski, Tacchetti, Raposo, Santoro, Faulkner,
  G{\"{u}}l{\c{c}}ehre, Song, Ballard, Gilmer, Dahl, Vaswani, Allen, Nash,
  Langston, Dyer, Heess, Wierstra, Kohli, Botvinick, Vinyals, Li, and
  Pascanu]{battaglia2018relationalinductivebiases}
Peter~W. Battaglia, Jessica~B. Hamrick, Victor Bapst, Alvaro
  Sanchez{-}Gonzalez, Vin{\'{\i}}cius~Flores Zambaldi, Mateusz Malinowski,
  Andrea Tacchetti, David Raposo, Adam Santoro, Ryan Faulkner, {\c{C}}aglar
  G{\"{u}}l{\c{c}}ehre, H.~Francis Song, Andrew~J. Ballard, Justin Gilmer,
  George~E. Dahl, Ashish Vaswani, Kelsey~R. Allen, Charles Nash, Victoria
  Langston, Chris Dyer, Nicolas Heess, Daan Wierstra, Pushmeet Kohli,
  Matthew~M. Botvinick, Oriol Vinyals, Yujia Li, and Razvan Pascanu.
\newblock Relational inductive biases, deep learning, and graph networks.
\newblock \emph{CoRR}, abs/1806.01261, 2018.
\newblock URL \url{http://arxiv.org/abs/1806.01261}.

\bibitem[Bronstein et~al.(2021)Bronstein, Bruna, Cohen, and
  Velickovic]{bronstein2021GeometricDeepLearning}
Michael~M. Bronstein, Joan Bruna, Taco Cohen, and Petar Velickovic.
\newblock Geometric deep learning: Grids, groups, graphs, geodesics, and
  gauges.
\newblock \emph{CoRR}, abs/2104.13478, 2021.
\newblock URL \url{https://arxiv.org/abs/2104.13478}.

\bibitem[Chaari et~al.(2022)Chaari, Gharsallaoui, Akda{\u{g}}, and
  Rekik]{chaari2022multigraph}
Nada Chaari, Mohammed~Amine Gharsallaoui, Hatice~Camg{\"o}z Akda{\u{g}}, and
  Islem Rekik.
\newblock Multigraph classification using learnable integration network with
  application to gender fingerprinting.
\newblock \emph{Neural Networks}, 151:\penalty0 250--263, 2022.

\bibitem[Chen et~al.(2020)Chen, Lin, Li, Li, Zhou, and
  Sun]{Chen_Lin_Li_Li_Zhou_Sun_2020}
Deli Chen, Yankai Lin, Wei Li, Peng Li, Jie Zhou, and Xu~Sun.
\newblock Measuring and relieving the over-smoothing problem for graph neural
  networks from the topological view.
\newblock \emph{Proceedings of the AAAI Conference on Artificial Intelligence},
  34\penalty0 (04):\penalty0 3438--3445, Apr. 2020.
\newblock \doi{10.1609/aaai.v34i04.5747}.
\newblock URL \url{https://ojs.aaai.org/index.php/AAAI/article/view/5747}.

\bibitem[Dash et~al.(2019)Dash, Shakyawar, Sharma, and Kaushik]{dash2019big}
Sabyasachi Dash, Sushil~Kumar Shakyawar, Mohit Sharma, and Sandeep Kaushik.
\newblock Big data in healthcare: management, analysis and future prospects.
\newblock \emph{Journal of Big Data}, 6\penalty0 (1):\penalty0 1--25, 2019.

\bibitem[Ding et~al.(2022)Ding, Xu, Tong, and Liu]{ding2022data}
Kaize Ding, Zhe Xu, Hanghang Tong, and Huan Liu.
\newblock Data augmentation for deep graph learning: A survey.
\newblock \emph{arXiv preprint arXiv:2202.08235}, 2022.

\bibitem[Fey et~al.(2018)Fey, Lenssen, Weichert, and
  M{\"u}ller]{fey2018splinecnn}
Matthias Fey, Jan~Eric Lenssen, Frank Weichert, and Heinrich M{\"u}ller.
\newblock Splinecnn: Fast geometric deep learning with continuous b-spline
  kernels.
\newblock In \emph{Proceedings of the IEEE conference on computer vision and
  pattern recognition}, pages 869--877, 2018.

\bibitem[Gilmer et~al.(2017)Gilmer, Schoenholz, Riley, Vinyals, and
  Dahl]{gilmer2017neural}
Justin Gilmer, Samuel~S Schoenholz, Patrick~F Riley, Oriol Vinyals, and
  George~E Dahl.
\newblock Neural message passing for quantum chemistry.
\newblock In \emph{International conference on machine learning}, pages
  1263--1272. PMLR, 2017.

\bibitem[Hong et~al.(2021)Hong, Wang, Han, and Ji]{hong2021spatiotemporal}
Genxuan Hong, Zhanquan Wang, Taoli Han, and Hengming Ji.
\newblock Spatiotemporal multi-graph convolutional network for taxi demand
  prediction.
\newblock In \emph{2021 11th International Conference on Information Science
  and Technology (ICIST)}, pages 242--250. IEEE, 2021.

\bibitem[Iglesias et~al.(2011)Iglesias, Liu, Thompson, and
  Tu]{iglesias2011robust}
Juan~Eugenio Iglesias, Cheng-Yi Liu, Paul~M Thompson, and Zhuowen Tu.
\newblock Robust brain extraction across datasets and comparison with publicly
  available methods.
\newblock \emph{IEEE Transactions on Medical Imaging}, 30\penalty0
  (9):\penalty0 1617--1634, 2011.

\bibitem[Kazi et~al.(2019)Kazi, Shekarforoush, Arvind~Krishna, Burwinkel,
  Vivar, Wiestler, Kort{\"u}m, Ahmadi, Albarqouni, and Navab]{kazi2019graph}
Anees Kazi, Shayan Shekarforoush, S~Arvind~Krishna, Hendrik Burwinkel, Gerome
  Vivar, Benedict Wiestler, Karsten Kort{\"u}m, Seyed-Ahmad Ahmadi, Shadi
  Albarqouni, and Nassir Navab.
\newblock Graph convolution based attention model for personalized disease
  prediction.
\newblock In \emph{International Conference on Medical Image Computing and
  Computer-Assisted Intervention}, pages 122--130. Springer, 2019.

\bibitem[Kim et~al.(2021)Kim, Ye, and Kim]{kim2021learning}
Byung-Hoon Kim, Jong~Chul Ye, and Jae-Jin Kim.
\newblock Learning dynamic graph representation of brain connectome with
  spatio-temporal attention.
\newblock \emph{Advances in Neural Information Processing Systems},
  34:\penalty0 4314--4327, 2021.

\bibitem[Kipf and Welling(2016)]{kipf2016semi}
Thomas~N Kipf and Max Welling.
\newblock Semi-supervised classification with graph convolutional networks.
\newblock \emph{arXiv preprint arXiv:1609.02907}, 2016.

\bibitem[LaMontagne et~al.(2019)LaMontagne, Benzinger, Morris, Keefe, Hornbeck,
  Xiong, Grant, Hassenstab, Moulder, Vlassenko, et~al.]{lamontagne2019oasis}
Pamela~J LaMontagne, Tammie~LS Benzinger, John~C Morris, Sarah Keefe, Russ
  Hornbeck, Chengjie Xiong, Elizabeth Grant, Jason Hassenstab, Krista Moulder,
  Andrei~G Vlassenko, et~al.
\newblock Oasis-3: longitudinal neuroimaging, clinical, and cognitive dataset
  for normal aging and alzheimer disease.
\newblock \emph{MedRxiv}, 2019.

\bibitem[Li(2020)]{li_2020}
Chuan Li.
\newblock Openai's gpt-3 language model: A technical overview, Sep 2020.
\newblock URL \url{https://lambdalabs.com/blog/demystifying-gpt-3/}.

\bibitem[Li et~al.(2017)Li, Cai, and He]{li2017learning}
Junying Li, Deng Cai, and Xiaofei He.
\newblock Learning graph-level representation for drug discovery.
\newblock \emph{arXiv preprint arXiv:1709.03741}, 2017.

\bibitem[Miller et~al.(2016)Miller, Alfaro-Almagro, Bangerter, Thomas, Yacoub,
  Xu, Bartsch, Jbabdi, Sotiropoulos, Andersson, Griffanti, Douaud, Okell,
  Weale, Dragonu, Garratt, Hudson, Collins, Jenkinson, Matthews, and
  Smith]{miller2016}
Karla~L Miller, Fidel Alfaro-Almagro, Neal~K Bangerter, David~L Thomas, Essa
  Yacoub, Junqian Xu, Andreas~J Bartsch, Saad Jbabdi, Stamatios~N Sotiropoulos,
  Jesper L~R Andersson, Ludovica Griffanti, Gwena{\"{e}}lle Douaud, Thomas~W
  Okell, Peter Weale, Iulius Dragonu, Steve Garratt, Sarah Hudson, Rory
  Collins, Mark Jenkinson, Paul~M Matthews, and Stephen~M Smith.
\newblock Multimodal population brain imaging in the {UK Biobank} prospective
  epidemiological study.
\newblock \emph{Nature Neuroscience}, 19\penalty0 (11):\penalty0 1523--1536,
  2016.
\newblock ISSN 1097-6256.
\newblock \doi{10.1038/nn.4393}.

\bibitem[Morris et~al.(2019)Morris, Ritzert, Fey, Hamilton, Lenssen, Rattan,
  and Grohe]{morris2019weisfeiler}
Christopher Morris, Martin Ritzert, Matthias Fey, William~L Hamilton, Jan~Eric
  Lenssen, Gaurav Rattan, and Martin Grohe.
\newblock Weisfeiler and leman go neural: Higher-order graph neural networks.
\newblock In \emph{Proceedings of the AAAI conference on artificial
  intelligence}, volume~33, pages 4602--4609, 2019.

\bibitem[Morris(1991)]{morris1991clinical}
John~C Morris.
\newblock The clinical dementia rating (cdr): Current version and.
\newblock \emph{Young}, 41:\penalty0 1588--1592, 1991.

\bibitem[Patenaude et~al.(2011)Patenaude, Smith, Kennedy, and
  Jenkinson]{patenaude2011bayesian}
Brian Patenaude, Stephen~M Smith, David~N Kennedy, and Mark Jenkinson.
\newblock A bayesian model of shape and appearance for subcortical brain
  segmentation.
\newblock \emph{Neuroimage}, 56\penalty0 (3):\penalty0 907--922, 2011.

\bibitem[Rong et~al.(2019)Rong, Huang, Xu, and Huang]{rong2019dropedge}
Yu~Rong, Wenbing Huang, Tingyang Xu, and Junzhou Huang.
\newblock Dropedge: Towards deep graph convolutional networks on node
  classification.
\newblock \emph{arXiv preprint arXiv:1907.10903}, 2019.

\bibitem[Rusu et~al.(2008)Rusu, Marton, Blodow, and Beetz]{rusu2008learning}
Radu~Bogdan Rusu, Zoltan~Csaba Marton, Nico Blodow, and Michael Beetz.
\newblock Learning informative point classes for the acquisition of object
  model maps.
\newblock In \emph{2008 10th International Conference on Control, Automation,
  Robotics and Vision}, pages 643--650. IEEE, 2008.

\bibitem[Rusu et~al.(2009)Rusu, Blodow, and Beetz]{rusu2009fast}
Radu~Bogdan Rusu, Nico Blodow, and Michael Beetz.
\newblock Fast point feature histograms (fpfh) for 3d registration.
\newblock In \emph{2009 IEEE international conference on robotics and
  automation}, pages 3212--3217. IEEE, 2009.

\bibitem[Sarasua et~al.(2022)Sarasua, P{\"o}lsterl, and
  Wachinger]{sarasua2022hippocampal}
Ignacio Sarasua, Sebastian P{\"o}lsterl, and Christian Wachinger.
\newblock Hippocampal representations for deep learning on alzheimer’s
  disease.
\newblock \emph{Scientific reports}, 12\penalty0 (1):\penalty0 1--13, 2022.

\bibitem[Shafto et~al.(2014)Shafto, Tyler, Dixon, Taylor, Rowe, Cusack, Calder,
  Marslen-Wilson, Duncan, Dalgleish, et~al.]{shafto2014cambridge}
Meredith~A. Shafto, Lorraine~K. Tyler, Marie Dixon, Jason~R. Taylor, James~B.
  Rowe, Rhodri Cusack, Andrew~J. Calder, William~D. Marslen-Wilson, John
  Duncan, Tim Dalgleish, et~al.
\newblock The {C}ambridge {C}entre for {A}geing and {N}euroscience ({Cam-CAN})
  study protocol: a cross-sectional, lifespan, multidisciplinary examination of
  healthy cognitive ageing.
\newblock \emph{BMC Neurology}, 14\penalty0 (1):\penalty0 204, 2014.

\bibitem[Shlomi et~al.(2020)Shlomi, Battaglia, and Vlimant]{shlomi2020graph}
Jonathan Shlomi, Peter Battaglia, and Jean-Roch Vlimant.
\newblock Graph neural networks in particle physics.
\newblock \emph{Machine Learning: Science and Technology}, 2\penalty0
  (2):\penalty0 021001, 2020.

\bibitem[Strubell et~al.(2019)Strubell, Ganesh, and
  McCallum]{strubell2019energy}
Emma Strubell, Ananya Ganesh, and Andrew McCallum.
\newblock Energy and policy considerations for deep learning in nlp.
\newblock \emph{arXiv preprint arXiv:1906.02243}, 2019.

\bibitem[Sudlow et~al.(2015)Sudlow, Gallacher, Allen, Beral, Burton, Danesh,
  Downey, Elliott, Green, Landray, et~al.]{sudlow2015uk}
Cathie Sudlow, John Gallacher, Naomi Allen, Valerie Beral, Paul Burton, John
  Danesh, Paul Downey, Paul Elliott, Jane Green, Martin Landray, et~al.
\newblock {UK Biobank}: an open access resource for identifying the causes of a
  wide range of complex diseases of middle and old age.
\newblock \emph{PLoS medicine}, 12\penalty0 (3):\penalty0 e1001779, 2015.

\bibitem[Taylor et~al.(2017)Taylor, Williams, Cusack, Auer, Shafto, Dixon,
  Tyler, Henson, et~al.]{taylor2017cambridge}
Jason~R. Taylor, Nitin Williams, Rhodri Cusack, Tibor Auer, Meredith~A. Shafto,
  Marie Dixon, Lorraine~K. Tyler, Richard~N. Henson, et~al.
\newblock The {C}ambridge {C}entre for {A}geing and {N}euroscience ({Cam-CAN})
  data repository: structural and functional {MRI}, {MEG}, and cognitive data
  from a cross-sectional adult lifespan sample.
\newblock \emph{NeuroImage}, 144:\penalty0 262--269, 2017.
\newblock \doi{https://doi.org/10.1016/j.neuroimage.2015.09.018}.

\bibitem[Tustison et~al.(2010)Tustison, Avants, Cook, Zheng, Egan, Yushkevich,
  and Gee]{tustison2010n4itk}
Nicholas~J Tustison, Brian~B Avants, Philip~A Cook, Yuanjie Zheng, Alexander
  Egan, Paul~A Yushkevich, and James~C Gee.
\newblock {N4ITK}: improved {N3} bias correction.
\newblock \emph{IEEE Transactions on Medical Imaging}, 29\penalty0
  (6):\penalty0 1310--1320, 2010.

\bibitem[Wang et~al.(2021)Wang, Chen, Zhang, Gong, Kumar, and
  Wei]{wang2021multi}
Wei Wang, Junyang Chen, Yushu Zhang, Zhiguo Gong, Neeraj Kumar, and Wei Wei.
\newblock A multi-graph convolutional network framework for tourist flow
  prediction.
\newblock \emph{ACM Transactions on Internet Technology (TOIT)}, 21\penalty0
  (4):\penalty0 1--13, 2021.

\bibitem[Wiggers(2020)]{wiggers_2020}
Kyle Wiggers.
\newblock Openai's massive gpt-3 model is impressive, but size isn't
  everything, Jun 2020.
\newblock URL
  \url{https://venturebeat.com/ai/ai-machine-learning-openai-gpt-3-size-isnt-everything/}.

\bibitem[Wu et~al.(2020)Wu, Pan, Chen, Long, Zhang, and
  Philip]{wu2020comprehensive}
Zonghan Wu, Shirui Pan, Fengwen Chen, Guodong Long, Chengqi Zhang, and S~Yu
  Philip.
\newblock A comprehensive survey on graph neural networks.
\newblock \emph{IEEE transactions on neural networks and learning systems},
  32\penalty0 (1):\penalty0 4--24, 2020.

\bibitem[Xu et~al.(2018)Xu, Hu, Leskovec, and Jegelka]{xu2018powerful}
Keyulu Xu, Weihua Hu, Jure Leskovec, and Stefanie Jegelka.
\newblock How powerful are graph neural networks?
\newblock \emph{arXiv preprint arXiv:1810.00826}, 2018.

\bibitem[Zhou et~al.(2020{\natexlab{a}})Zhou, Shen, and Xuan]{zhou2020data}
Jiajun Zhou, Jie Shen, and Qi~Xuan.
\newblock Data augmentation for graph classification.
\newblock In \emph{Proceedings of the 29th ACM International Conference on
  Information \& Knowledge Management}, pages 2341--2344, 2020{\natexlab{a}}.

\bibitem[Zhou et~al.(2020{\natexlab{b}})Zhou, Cui, Hu, Zhang, Yang, Liu, Wang,
  Li, and Sun]{zhou2020graph}
Jie Zhou, Ganqu Cui, Shengding Hu, Zhengyan Zhang, Cheng Yang, Zhiyuan Liu,
  Lifeng Wang, Changcheng Li, and Maosong Sun.
\newblock Graph neural networks: A review of methods and applications.
\newblock \emph{AI Open}, 1:\penalty0 57--81, 2020{\natexlab{b}}.

\end{thebibliography}

\appendix

\end{document}